\pdfoutput=1

\documentclass[11pt,letterpaper]{article}

\usepackage[]{acl}

\usepackage{times}
\usepackage{latexsym}
\usepackage{amsmath}
\usepackage{amssymb}

\usepackage[T1]{fontenc}
\usepackage[utf8]{inputenc}

\usepackage{microtype}

\usepackage{graphicx}

\title{What \textit{Drives} the Use of Metaphorical Language? \\ Negative Insights from Abstractness, Affect, Discourse Coherence and Contextualized Word Representations}

\author{Prisca Piccirilli and Sabine Schulte im Walde \\
  Institute for Natural Language Processing, University of Stuttgart, Germany \\
  \texttt{\{prisca.piccirilli,schulte\}@ims.uni-stuttgart.de} \\}

\begin{document}
\maketitle

\begin{abstract}
Given a specific discourse, which discourse properties trigger the use of metaphorical language, rather than using literal alternatives? For example, what drives people to say \textit{grasp the meaning} rather than \textit{understand the meaning} within a specific context? Many {\small NLP} approaches to metaphorical language rely on cognitive and (psycho-)linguistic insights and have successfully defined models of discourse coherence, abstractness and affect. In this work, we build five simple models
relying on established cognitive and linguistic properties -- frequency, abstractness, affect, discourse coherence and contextualized word representations -- to predict the use of a metaphorical vs. synonymous literal expression in context. By comparing the models' outputs to human judgments, our study indicates that our selected properties are not sufficient to systematically explain metaphorical vs.~literal language choices.

\end{abstract}

\section{Introduction}
\label{sec:intro}

Metaphors are "not just nice", but represent a "necessary" element of everyday thought and communication \citep[i.a.]{Ortony:1975,Lakoff-Johnson:1980,vandenBroek:1981,Schaeffner:2004},
and are ubiquitous in natural language text corpora \citep[i.a.]{Gedigian-etal:2006,Shutova-Teufel:2010,Steen-etal:2010}.
From the perspective of natural language processing ({\small NLP}), automatic approaches to metaphor processing are therefore important for any task that requires natural language understanding, and {\small NLP} has been concerned with the detection \citep[i.a.]{Koeper-SchulteImWalde:2016b,Alnafesah-etal:2020,Ehren-etal:2020,Dankers-etal:2020}, the interpretation \citep[i.a.]{Shutova:2010,Bizzoni-Lappin:2018,Mao-etal:2018} and, 
the generation \citep{Stowe-etal:2021,Zhou-etal:2021} of metaphors.\footnote{See \citet{Tong-etal:2021} for a systematic, comprehensive review and discussion of the most recent metaphor processing systems and datasets.}

As to our knowledge, however, no study so far has raised the question of {\small WHY} a metaphorical expression is used within a specific discourse, rather than an equally plausible literal alternative.
For example, consider the discourse in table~\ref{tab:example-pair},
where both the metaphorical expression \textit{grasp the meaning} and its synonymous literal alternative \textit{understand the meaning} seem equally acceptable \citep{Piccirilli-SchulteImWalde:2021,Piccirilli-SchulteImWalde:2022}.
What are the factors priming for one use or the other? Are there cues within the discourse which influence the selection of one usage over the other? To which extent can computational approaches based on discourse properties model human behavior regarding the choice between synonymous metaphorical vs. literal language usage?

\begin{table*}[th!]
\begin{tabular}{p{0.97\linewidth}}
\hline
\footnotesize
This wasn't just a play on words, rather it was a demand that they should 'maintain a consistency between their words and their actions'. But I agree, that still does not absolve them from the need to speak truth to power. In our times when people spend so much time with TV and the internet, do they have the interest and time to read poetry? Many people believe that it is difficult to read poetry. Can everyone [\textit{\textcolor{red}{grasp}} / \textit{\textcolor{blue}{understand} the meaning}] of a good poem, or is a skill necessary?\\
\hline
\end{tabular}
\captionsetup{font=footnotesize}
\caption{Example of a discourse from the dataset introduced by \citet{Piccirilli-SchulteImWalde:2021}. 
Both the literal expression \textit{\textcolor{blue}{understand} meaning} and its metaphorical counterpart \textit{\textcolor{red}{grasp} meaning} seem equally acceptable in the discourse.}
\label{tab:example-pair}
\vspace{-2mm}
\end{table*}

According to psycholinguistics and computational linguistics research, the processing of words is a function of their \textbf{frequency of occurrence} in the language \citep[i.a.]{vanJaarsveld-Rattink:1988,Wittmann-etal:2017}. Is the choice between a metaphorical and a literal expression therefore just an effect of frequency?
In contrast, conceptual metaphor theory establishes metaphorical language as a figurative device for transferring knowledge from a concrete domain to a more abstract domain \cite{Lakoff-Johnson:1980}, and the hypothesis that metaphorical usages correlate with the \textbf{abstractness of the context} has been supported in numerous {\small NLP} studies on automatic metaphor identification
\cite{Turney-etal:2011,Tsvetkov-etal:2013,Koeper-SchulteImWalde:2016b,Alnafesah-etal:2020,Hall-etal:2020}. 
\textbf{Affect} has also been explored with regard to metaphoricity. 
Not only has metaphorical language been found to carry a stronger emotional load than literal language \cite{Blanchette-Dunbar:2001,Crawford:2009}, but metaphorical words and sentences are also judged to be "more emotionally engaging" than their synonymous literal paraphrases \cite{Citron-Goldberg:2014,Mohammad-etal:2016}, and
informing {\small NLP} models with \textbf{emotion} features has proven useful for metaphor detection \cite{Garget-Barnden:2015,Koeper-Schulte-im-walde:2018,Dankers-etal:2019}.
When analyzing metaphorical discourse features \cite{Glucksberg:1989,Steen:2004} and the interactions of \textbf{discourse coherence} with the contextual salience of metaphorical vs.~literal expressions \cite{Inhoff-etal:1984,Gibbs:1989, Giora:1997,Giora-Fein:1999,Gibbs:2002,Koevecses:2009},
findings are directly connected with the theory of discourse cohesion, i.e., the principle that discourse should be a "group of collocated, structured, and coherent sentences" \cite{Halliday-Hasan:1976,Jurafsky-Martin:2019}.
Models relying on the coherence of lexical semantic discourse structures have therefore been successful in identifying metaphors \cite{Sporleder-Li:2009, Bogdanova:2010,Mesgar-Strube:2016,Dankers-etal:2020}. 
From a yet different perspective, Transformer-based pretrained language models ({\small T-PLMs}) \cite{Vaswani-etal:2017}, pre-trained on the language modeling task, are able to predict masked items in context and produce \textbf{contextual embeddings} accounting for both left and right contexts, and resulting in word representations that are dynamically informed by the surrounding words.

The rich previous interdisciplinary research on figurative language seems to agree on tight interactions between metaphorical language detection and properties of the respective discourses, i.e., \textbf{cognitive aspects} (abstractness and affect), \textbf{discourse coherence} and  \textbf{contextual properties}.  
Do these discourse properties indeed trigger the use of a metaphorical vs. its synonymous counterpart? We address this question by exploring five simple discourse-based models inspired by the above-mentioned properties, namely frequency, abstractness, affect, discourse coherence and contextualized word representations. 
We approach the question within two studies. 
First, we explore discourse features of metaphorical usages occurring in natural language by applying the cognitive and linguistically-inspired models to existing discourses from the English corpus \textit{ukWaC} \cite{Baroni-etal:2009}. 
Then, we zoom into the models' predictions and evaluate them against human preferences (i.e., annotations) for metaphorical vs. literal language within the same discourses. 
By modeling the prediction for synonymous metaphorical vs. literal expressions motivated by the above discourse perspectives, we gain insight into human behavior for metaphorical vs. literal languages choices.

\section{Dataset}
\label{sec:data}

Research on figurative language has produced impressive resources on the metaphoricity of lexical items. However, these resources have limitations regarding the specific task we are addressing in this work: (i) the context in which words are metaphorically used is not large enough, providing human judgments only on the \textit{word-level} \cite{Steen-etal:2010} or on the \textit{sentence-level} \cite{Stefanowitsch:2008,Shutova-Teufel:2010,Mohammad-etal:2016}, (ii) the target words or expressions are \textit{ambiguous}, i.e., they may have both a metaphorical and a literal sense \cite{Tsvetkov-etal:2013, Mohler-etal:2016}, (iii) they are \textit{extended} metaphors \cite{Gibbs:2006,Martin:2008}, (iv) the paraphrases to metaphorically-used words are automatically generated and no manual annotations were provided to evaluate the outputs \citep[i.a.]{Bollegala-Shutova:2013,Bizzoni-Lappin:2018}.

We therefore use our recently released dataset specifically designed to investigate the choice of metaphorical vs. literal expressions in context \cite{Piccirilli-SchulteImWalde:2021,Piccirilli-SchulteImWalde:2022}. It contains a total of 1,000 discourses of five to six sentences (98 words on average), in which the final sentence of each discourse contains either a metaphorical expression or its literal alternative from a pair of synonymous subject--verb {\small (SV)} or verb--object {\small (VO)} expressions. Table \ref{tab:example-pair} presents an example for the {\small VO} pair \textit{grasp/understand meaning}. The overall 50 pairs of English expressions were selected from \citet{Shutova:2010} and \citet{Mohammad-etal:2016}, and for each of the pairs, we extracted 20 discourses from the \textit{ukWaC} \cite{Baroni-etal:2009}, 10 of which containing the metaphorical usage (e.g., \textit{grasp}) and 10 of which containing the literal paraphrase (e.g., \textit{understand}). 
To gain insight into human preferences between the selected pairs of metaphorical vs. literal expressions, we also collected crowdsourced human judgments for these 1,000 discourses, asking annotators to choose which expression they favored given the preceding discourse. 
This dataset is therefore optimal for the task at hand, as it provides (i) synonymous metaphorical vs. literal expressions, (ii) at the \textit{discourse-level}, (iii) manually annotated.
In the present work, we make use of the 1,000 discourses containing the original expressions in the \textit{ukWaC} 
and the annotators' choices for metaphorical vs. literal usages within a subset of 287 discourses, where 70\% or more annotators agreed on the preference (metaphorical or literal).

\section{Models and Experimental Setup}
\label{sec:models}

\subsection{Prediction models}

We approach the task of predicting the use of metaphorical vs.~literal expressions as a \textit{prompting task}: given an input prompt\footnote{The input prompt is a \textit{prefix prompt} for all models except for {\small BERT}, whose input is a \textit{cloze prompt.}}, the models predict whether the missing span should be the metaphorical or the literal expression. We apply five models relying on discourse properties, where each model approaches the task from a different perspective, to give us insight on which discourse features influence the metaphorical vs.~literal selection.

\paragraph{Frequency approach}
When given the choice between a metaphorical expression and its synonymous counterpart, do we tend to favor the most frequent usage? 
\textbf{Baseline ({\small Freq.}):}
Our baseline relies on the occurrences of the {\small SV} and {\small OV} tuples in the original \textit{ukWaC} corpus: the model receives the prefix prompt as input, and always outputs the most frequent expression of the pair. 

\paragraph{Cognitively-inspired approaches}
The cognitive interaction between abstractness/affect and metaphorical language raises the question: to which extent (i)~a more abstract discourse and (ii)~a more emotionally-loaded discourse favor a metaphorical usage? 
\textbf{Abstractness ({\small Abstr.}):}
We measure the abstractness of a discourse preceding the target expression within four settings, based on the norms from \citet{Brysbaert-etal:2014}. We assign abstractness scores to \textit{all} words ({\small Abstr.all}), only \textit{nouns} ({\small Abstr.n}), only \textit{verbs} ({\small Abstr.v}) or only \textit{adjectives} ({\small Abstr.adj}). We then obtain an overall rating of abstractness for each discourse by computing the median of the respective lexical items' abstractness scores. For each setting, we use as threshold the abstractness median of the respective part-of-speech class, and the model predicts the metaphorical expression if the overall abstractness score of the discourse is below that threshold (i.e., more abstract/less concrete), and the literal counterpart if above (i.e., less abstract/more concrete). 
\textbf{Emotionality ({\small Emo.}):}
We build a model predicting the target expression based on the emotionality of the preceding discourse, which we represent using the English emotion lexicon from \citet{Buechel-etal:2020}. 
Each lexical item from the preceding discourse is assigned an emotionality score, and the median represents the overall emotionality score of the discourse. Appendix \ref{app:emotionality} provides details on the emotionality score.

\paragraph{Discourse coherence approach}
Is the choice of an expression from a synonymous pair driven by the semantic relatedness between the components of that expression and the lexical items in the surrounding context? We adapt the 
\textbf{Lexical Coherence Graph ({\small LCG})} introduced by \citet{Mesgar-Strube:2016} to measure the semantic relatedness between the words in the preceding context and the target expression contained in the final discourse sentence: we compute the cosine scores for the two components in the {\small SV/VO} expressions (i.e., the verb and its argument) and each word in the preceding discourse, relying on contextualized {\small BERT} embeddings \cite{Devlin-etal:2019}. The output is a graph connecting the preceding context to both the metaphorical expression and its synonymous alternative; edge weights are represented by the average of the respecting cosine values. The expression -- metaphorical or literal -- with the maximum weight (i.e., the largest average cosine score) is selected. Appendix \ref{app:lcg} provides details on the {\small LCG} score.

\paragraph{Contextualized discourse properties}
How do contextualized word representations prime for the use of a metaphorical vs.~literal preference?
Our question triggers a \textit{cloze-task} style \cite{Taylor:1953} and applies the \textbf{Pre-trained Language Model {\small BERT}} \cite{Devlin-etal:2019} in a zero-shot manner.
We give a \textit{cloze prompt} as input to the model, as in table \ref{tab:example-pair}: we mask the target expression, and the model selects the most probable answer amongst the two candidates (metaphorical or literal). We experiment with both {\small BERT\textsubscript{base}} and {\small BERT\textsubscript{large}}.

\subsection{Experimental Setup}
\label{sec:setup}
We compare our models' predictions against two gold standards.  
We first evaluate the models' outputs against the 1,000 discourses collected from \textit{ukWaC}, containing the balanced \textit{originally}-used metaphorical or literal expressions (\textbf{Orig. Data})  
to investigate which discourse aspects (abstractness, emotionality, coherence, word representations) might have primed the metaphorical vs.~literal selection. 
We expect the model predictions to provide insight into the features that influenced the \textit{original} speakers' preferences.
We then zoom into the \textit{annotated} version of the same data (\textbf{Anno. Data}), for which participants were asked which expression, metaphorical or literal, they favored in the given 287 discourses (cf. Section \ref{sec:data}). We analyze whether (dis-)agreements between humans are also reflected in the models' predictions. 

\section{Results and Analyses}
\label{sec:results}

\begin{figure}[t]
   \includegraphics[scale=0.57]{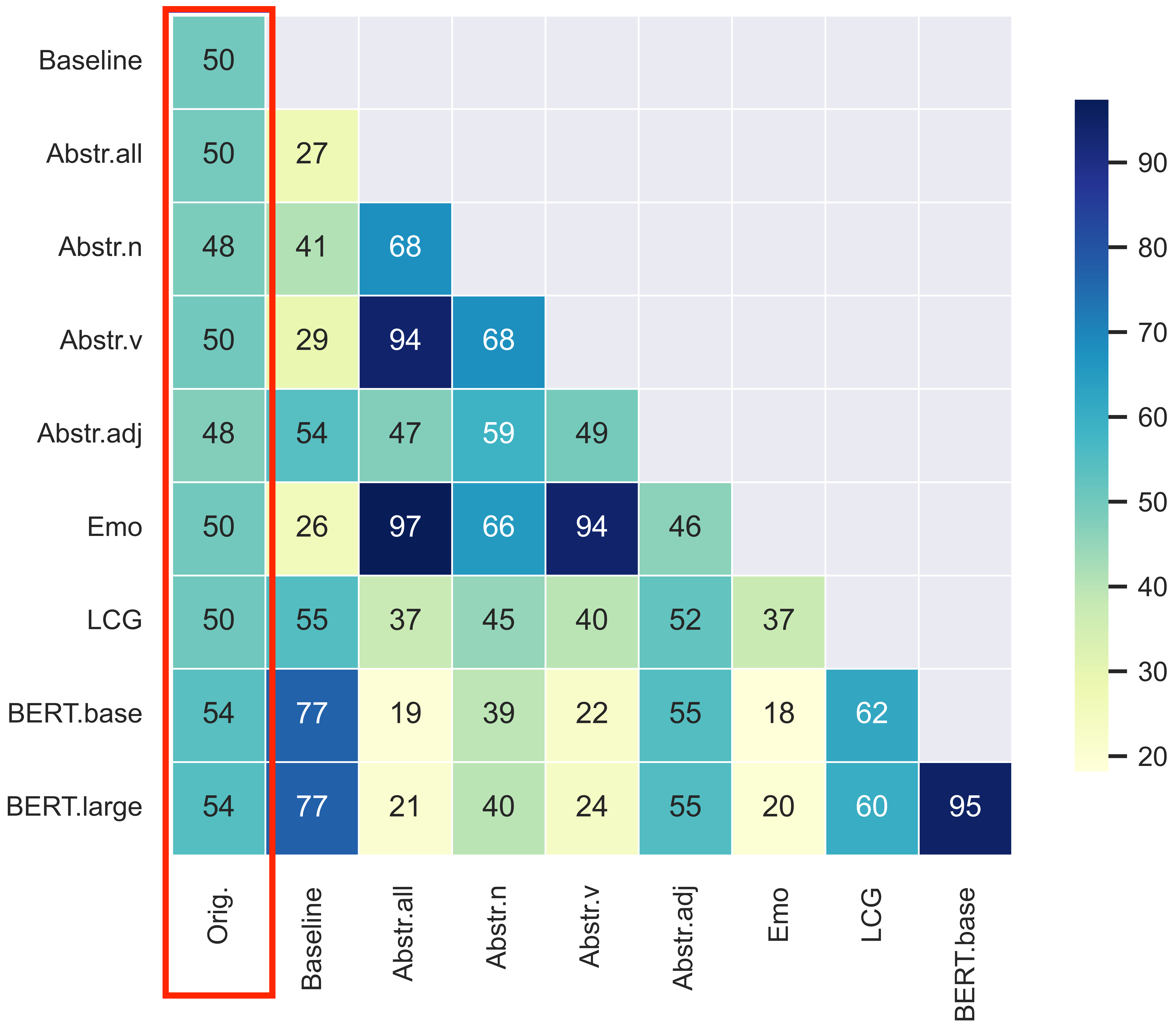}
    \captionsetup{font=footnotesize}
    \caption{Accuracy scores of each model with regard to the original data (1\textsuperscript{st} column in \textcolor{red}{red}) and percentages of overlapping output decisions between models.}
    \label{fig:contingency}
    \vspace{-3.5mm}
\end{figure}

We first analyze the models' predictions with regard to metaphorical vs. literal usages in the \textit{originally}-extracted data (\textbf{Orig. Data}), to address the question: do cognitive and linguistically-inspired models reflect metaphorical language usage encountered in natural language corpora? 
The first column of figure \ref{fig:contingency} presents the accuracy score of each model with regard to Orig. Data.
All models reach around 50\% of accuracy; however, they behave very differently regarding their individual predictions.\footnote{Appendix \ref{app-subsec:eval-orig} presents further prediction differences.}
Figure \ref{fig:contingency} presents the percentages of overlapping output decisions between our five models, revealing interesting insights:
we observe 77\% overlapping predictions between the {\small PLMs} with the frequency baseline, suggesting that the majority of {\small PLM} predictions is frequency-driven, which is not the case for most abstractness settings and emotionality. 
Emotionality itself correlates with abstractness in all settings but the one where we consider only adjectives, which is surprising as adjectives tend to carry a lot of emotions \cite{Mohammad-Turney:2010, Bostan-Klinger:2019}.
The overlap between {\small LCG} and {\small BERT} reaches 62\%, whereas its correlations with abstractness and emotionality are rather low, suggesting that decisions based on abstractness and emotionality are different to those based on word representations and semantic relatedness. At first sight, our different perspectives seem rather complementary, which is yet to be verified in future studies.

Overall, none of our models seems to reliably predict what is observed in natural language data. We thus cannot derive what might have triggered the original speakers to favor a metaphorical or literal expression over the other.

\begin{figure*}[ht]
    \centering
    \includegraphics[scale=0.6]{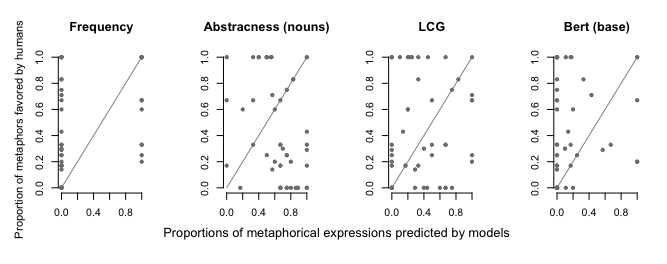}
    \vspace{-2mm}
    \caption{Proportions of predicted metaphorical expressions for each model, with regard to the proportions of these metaphorical expressions favored by annotators. Appendix \ref{app:eval-anno} shows the graphs for all models.}
    \label{fig:props}
    \vspace{-1.5mm}
\end{figure*}

We then compare our models' predictions to human behavior: are the proportions of metaphorical vs.~literal predictions from our models similar to human perceptions (\textbf{Anno. Data})?
Figure \ref{fig:props} presents the proportions of the metaphorical expressions predicted by four of the models in relation to the proportions of metaphorical usages favored by the human judges.\footnote{Appendix \ref{app:eval-anno} shows the graphs for all models.}
If humans and models were making similar decisions, all data points, each representing the proportion of metaphorical uses for a pair of expressions, would be on the regression line.
We make different observations depending on the model in question. As far as frequency is concerned, the model predicts either the metaphorical or the literal expression (100\% or 0\% respectively, in the graph). Many literal expressions that are favored by participants (low x-axis \%) seem to correlate with their higher occurrences in natural language corpora, e.g.,~\textit{make} vs.~\textit{throw remark} ({\small VO}).  
However, humans also favor many metaphorical expressions which are less frequently used, e.g.,~\textit{stir} vs.~\textit{cause excitement} ({\small VO}) and inversely, a few metaphorical expressions which are more frequent than their literal counterparts are not necessarily favored by people, e.g.,~\textit{poison} vs.~\textit{corrupt mind} ({\small VO}). Thus, frequency does not seem to be a systematic factor for metaphorical/literal choices.
Regarding abstractness, many expressions fall perfectly on the regression line, e.g.,~\textit{twist vs. misinterpret word} ({\small VO}), \textit{story grab vs. intrigue} ({\small SV}), suggesting at first some interactions between the expression preferences and the abstractness of the respective discourses. However, the numerous metaphorical expressions that are predicted by the model (80\%+) but not by humans, e.g.,~\textit{taste} vs.~\textit{experience freedom} ({\small VO}), \textit{factor shape} vs.~\textit{determine} ({\small SV}) do not confirm that hypothesis. 
Concerning the {\small LCG}, the picture is less clear. Many literal expressions are favored by both the model and humans, e.g., \textit{color} vs.~\textit{affect judgement} ({\small VO}), but many metaphorical usages are preferred by humans when the model predicts the literal ones, e.g.,~\textit{breathe} vs.~\textit{instill life} ({\small VO}). Therefore, metaphorical vs.~literal coherence does not seem either to be a determining factor for metaphorical vs.~literal preference, respectively, which does not support the context-salience hypothesis, where one would expect a metaphorical expression to be favored following a metaphorical discourse, and ditto for a literal expression/literal discourse \cite{Koevecses:2009}.
Finally, the overall low metaphorical predictions by the {\small PLM} are expressions for which humans provide high metaphorical proportions, such as \textit{abuse} vs.~\textit{drink alcohol}  ({\small VO}).
This suggests a potential lack of metaphorical language representation, in line with our observation concerning abstractness and emotionality. 

\section{Discussion and Future Work}
\label{sec:conclu}

Previous research has provided evidence for interactions between metaphorical language and discourse properties, namely frequency, abstractness, affect, discourse coherence and contextualized word representations, on the \textit{word-} and \textit{sentence-level}. In this work, we took a step further: we looked at the task of predicting the use of a metaphorical vs. synonymous literal expression and built models based on the above-mentioned features on the \textit{discourse-level}. 
Our findings show that these discourse properties do not seem to be indicative of metaphorical usage. 

We propose several directions for future work.
First of all, we considered discourses of around five sentences, but the decision on the context window might have an impact on the findings. Further work exploring the optimal size of preceding context would be interesting. 
Another promising direction might analyze {\small PLMs}' attention mechanisms \cite{Tenney-etal:2019,Clark-etal:2019} on the presented task, and also explore the extent to which modifying the attention of such models, i.e., fine-tuning \cite{Peters-etal:2019,Zhao-Bethard:2020}, improves their performance to mimic human preferences for metaphorical vs. literal usages. 
Finally, we have looked at five features; needless to say that exploring further discourse properties is necessary, such as co-reference, complexity, aptness, creativity, prototypicality, and the influence of genre and specific domains (e.g.,~religious/scientific texts).

\section{Conclusion} We suggested five simple models to investigate \textsc{why} humans choose to use a metaphorical expression in a specific discourse.
Regardless of the perspectives, our work demonstrates that a range of previously suggested salient discourse properties do not seem to influence preferences on the choice between synonymous metaphorical vs.~literal expressions. 
Our findings thus ask for a more nuanced approach to metaphorical language choices in {\small NLP}.

\section*{Acknowledgements}

We thank the anonymous reviewers for their useful and quality feedback.
This research was supported by the DFG Research Grant SCHU 2580/4-1 \textit{Multimodal Dimensions and Computational Applications of Abstractness}.

\bibliography{PriPicciBib}
\bibliographystyle{acl_natbib}

\clearpage
\appendix

\section{Models}

\subsection{Emotionality}
\label{app:emotionality}
\citet{Buechel-etal:2020} presented a new methodology to automatically generate lexicons for 91 languages comprising eight emotional variables: \textit{Valence, Arousal, Dominance} ({\small VAD}) as well as the five basic emotions \textit{Joy, Anger, Surprise, Fear, Surprise} ({\small BE5}) \cite{Ekman:1992}. 
As a source dataset, they used the English emotion lexicon from \citet{Warriner-etal:2013}, comprising about 14K entries in {\small VAD} format collected via crowdsourcing. 
They applied the {\small BE5} ratings from \citet{Buechel-Hahn:2018a} to convert the {\small VAD} ratings. Via their monolingual state-of-the-art multi-task feed-forward network \cite{Buechel-Hahn:2018b}, they projected ratings on these eight variables, resulting in an English lexicon containing 2M word type entries with very high correlation with human judgments (around 90\% for each variable). 

We are interested in the emotionality of lexical items, i.e., the emotional load that a term conveys, rather than the actual emotion a term refers to. 
We therefore use the {\small BE5} ratings of the English lexicon for our study. 
Out of the five scores that a term receives -- one for each emotion, we assume that its highest score is reflective of the "emotional load" of that term, i.e., how much emotion it conveys. For example, the lexical item "truth" obtained the scores 2.24, 1.46, 1.4, 1.49, 1.46 for Joy, Anger, Sadness, Fear and Surprise, respectively. 
In our experiments, the term "truth" is therefore attributed the score of 2.24 as its emotional load. 

\subsection{Lexical Coherence Graph}
\label{app:lcg}
\begin{figure*}[]
\centering
    \includegraphics[scale=0.7]{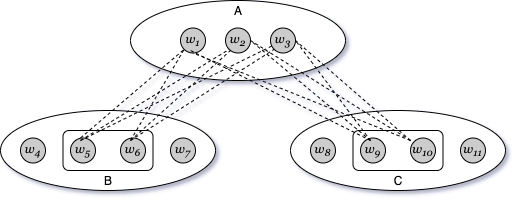}
    \caption{Discourse \textit{A} with three words \{\textit{$w_1,w_2,w_3$}\} and sentences \textit{B} and \textit{C} with four words, where \{\textit{$w_5,w_6$}\} are the two words composing the metaphorical expression, and \{\textit{$w_9,w_{10}$}\} are composing the literal paraphrase. Depending on the expression input ({\small SV} or {\small VO}), the respective subject or object is identical in \{\textit{$w_5,w_6$}\} and \{\textit{$w_9,w_{10}$}\}, as only the verb is used either as a metaphorical or a literal variant. The semantic relatedness between each word in \{\textit{$w_5,w_6$}\} and in  \{\textit{$w_9,w_{10}$}\} is computed with each word in \textit{A}.}
    \label{fig:my-model-step1}
\end{figure*}

\begin{figure*}[]
\centering
    \includegraphics[scale=0.7]{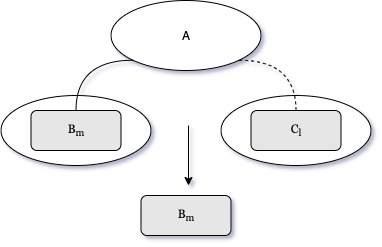}
    \caption{The word relation with the maximum weight (here \textit{$B_m$} as indicated by the plain line) represents a stronger connection with the preceding discourse \textit{A}.}
    \label{fig:my-model-step2}
\end{figure*}

Following \citet{Mesgar-Strube:2016}, we measure the semantic relatedness between words represented by their word embeddings, computing the cosine score between the two words of the expression ({\small SV} or {\small VO}) with each word of the preceding discourse. 
Consider \textit{$v_a$}, \textit{$v_b$}, \textit{$v_c$}, the word vectors for word \textit{a} in the preceding discourse \textit{A}, word \textit{b} in the metaphorical expression \textit{$B_m$} contained in the last sentence \textit{$B$} and word \textit{c} in the literal expression \textit{$C_l$} contained in the last sentence \textit{$C$}, respectively.
The cosine scores $\cos(v_a, v_b)$ and $\cos(v_a, v_c)$ between the two word vectors is a measure of semantic connectivity of the two words. The range of $\cos(v_a, v_b)$ and $\cos(v_a, v_c)$ is between $[-1, +1]$, showing how well the two words are semantically correlated.
Figure \ref{fig:my-model-step1} shows how relatedness is measured, and figure \ref{fig:my-model-step2} shows the output of the graph.

\section{Results}
\label{app:results}
\subsection{Evaluation: Predictions vs. Original Data}
\label{app-subsec:eval-orig}
\begin{figure}[h!]
    \includegraphics[scale=0.54]{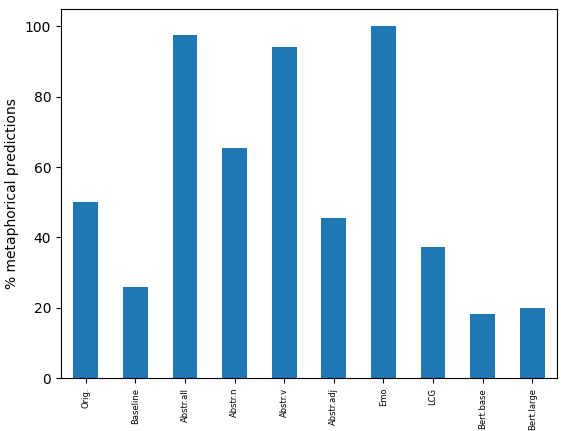}
    \caption{Percentages of metaphorical expressions predicted by each model.}
    \label{fig:props-met}
\end{figure}

Figure \ref{fig:props-met} presents the percentages of metaphorical expressions predicted by each model. As mentioned in section \ref{sec:results}, the accuracy scores of all models reach around 50\%, but they actually behave very differently with regard to performances for metaphorical vs.~literal predictions. 
Remember that the originally-collected discourses are perfectly balanced (cf. Section \ref{sec:data}), where half of the discourses contains metaphorical expressions and the other half contains literal expressions from synonymous pairs. As expected, the frequency baseline therefore reaches 50\% of accuracy. 
We note however that the literal usage of the pairs is the most frequent in the \textit{ukWaC} corpus (37/50 pairs), which leads the model to predict the literal counterpart in 74\% of the cases. 
The abstractness and emotionality models mostly select for the metaphorical usages (up to 100\% for Emo.), as the discourses are considered more abstract and more emotionally-loaded based on the respective norms. This suggests that original speakers did not perceive the degree of abstractness and emotionality of the discourse as triggers to favor one usage over the other. This aligns with the findings in \citet{Piccirilli-SchulteImWalde:2022}, who analyzed the relationship between the abstractness and emotionality of the preceding discourses with human preferences for metaphorical vs. literal expressions.
The {\small LCG} model favors the literal expressions as well (62\%), while both {\small BERT\textsubscript{base}} and {\small BERT\textsubscript{large}} predominantly predict the literal expressions (81.80\% and 80.10\%, respectively.). 

\onecolumn
\subsection{Evaluation: Predictions vs. Annotated Data}
\label{app:eval-anno}
\begin{figure}[h]
\centering
    \includegraphics[scale=0.55]{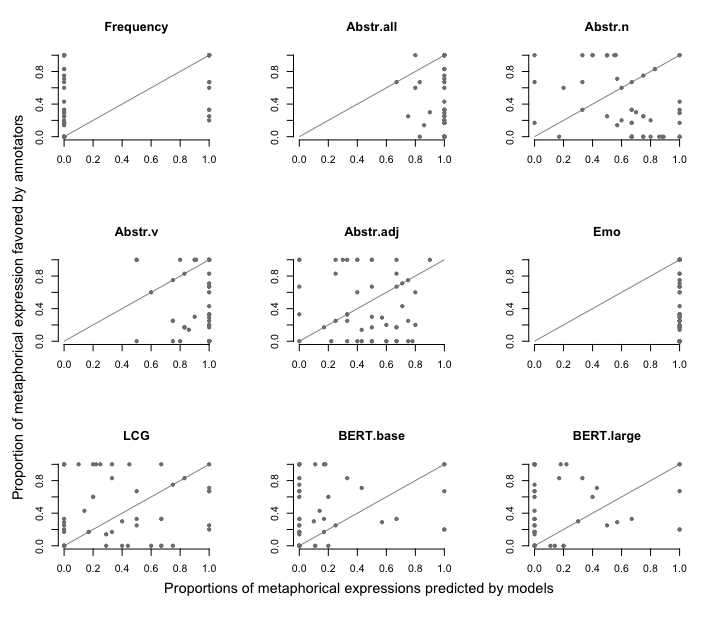}
    \caption{Proportions of the metaphorical expressions predicted by the models with regard to the proportions of these usages to be favored by the participants.}
    \label{fig:allprops}
\end{figure}

Figure \ref{fig:allprops} presents an overview of the proportion of expressions that are predicted metaphorically by the models with regard to the preferences of the same expressions by human annotators.
\vspace{.3cm}

\end{document}